\documentclass[10pt,twocolumn,letterpaper]{article}

\usepackage{iccv}              

\definecolor{iccvblue}{rgb}{0.21,0.49,0.74}
\usepackage[pagebackref,breaklinks,colorlinks,allcolors=iccvblue]{hyperref}
\usepackage{multirow}
\usepackage{colortbl} 
\def\paperID{5974} 
\def\confName{ICCV}
\def\confYear{2025}
\definecolor{red1}{RGB}{231,098,084}
\definecolor{blue1}{RGB}{114,188,213}
\definecolor{lightblue}{RGB}{173, 216, 230}
\title{FairGen: Enhancing Fairness in Text-to-Image Diffusion Models via Self-Discovering Latent Directions}

\author{
Yilei Jiang$^{1}$ \quad 
Wei-Hong Li$^{1,2}$\thanks{Corresponding author.} \quad 
Yiyuan Zhang$^{1}$ \quad 
Minghong Cai$^{1}$ \quad 
Xiangyu Yue$^{1,3}$\footnotemark[1] \\
$^{1}$MMLab, CUHK\quad 
$^{2}$University of Bristol\quad 
$^{3}$SHIAE, CUHK \\
}

\begin{document}
\maketitle
\begin{abstract}
While Diffusion Models (DM) exhibit remarkable performance across various image generative tasks, they nonetheless reflect the inherent bias presented in the training set. As DMs are now widely used in real-world applications, these biases could perpetuate a distorted worldview and hinder opportunities for minority groups. Existing methods on debiasing DMs usually requires model retraining with a human-crafted reference dataset or additional classifiers, which suffer from two major limitations: (1) collecting reference datasets causes expensive annotation cost; (2) the debiasing performance is heavily constrained by the quality of the reference dataset or the additional classifier. To address the above limitations, we propose FairGen, a plug-and-play method that learns attribute latent directions in a self-discovering manner, thus eliminating the reliance on such reference dataset. Specifically, FairGen consists of two parts: a set of attribute adapters and a distribution indicator. Each adapter in the set aims to learn an attribute latent direction, and is optimized via noise composition through a self-discovering process.
Then, the distribution indicator is multiplied by the set of adapters to guide the generation process towards the prescribed distribution. Our method enables debiasing multiple attributes in DMs simultaneously, while remaining lightweight and easily integrable with other DMs, eliminating the need for retraining. Extensive experiments on debiasing gender, racial, and their intersectional biases show that our method outperforms previous SOTA by a large margin. Our code is made publicly avaibable at \url{https://github.com/leigest519/FairGen}.
\end{abstract}    
\section{Introduction}
State-of-the-art Text-to-Image Diffusion Models (DMs) such as Stable Diffusion~\citep{rombach2022high}, DALL-E 3~\citep{ramesh2022hierarchical} and Imagen~\citep{saharia2022photorealistic} have demonstrated remarkable performance in generating high-quality images. With the rapid development of DMs, an increasing number of individuals and corporations are choosing to utilize them to serve their own purposes. For instance, Stable Diffusion v1.5 has been downloaded over 8 million times on Huggingface, and Midjourney is used by over a million users \citep{fatunde2022digital}. However, existing DMs have been found to generate biased content across various demographic factors, such as gender and race~\citep{luccioni2023stable}, which could have harmful effects on society when these models are implemented in real-world applications. 

\begin{figure}[t!]
\centering
    \includegraphics[width=0.95\linewidth]{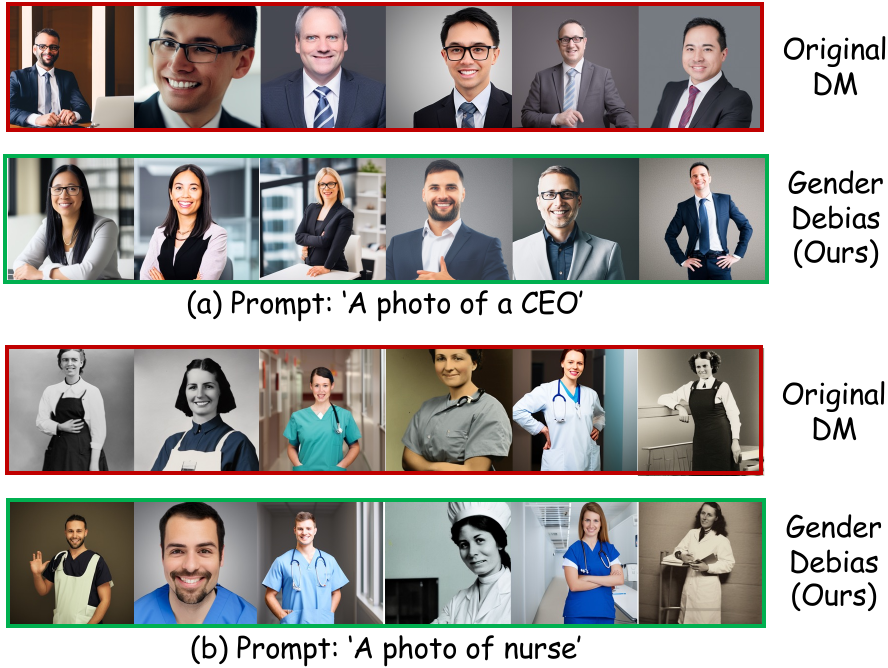} 
    \vspace{-0.3cm}
    \caption{Illustration of gender bias associated with different occupations in Stable Diffusion v2.1 and the debiased results for gender by our method. The leadership roles such as CEOs are biased towards male figures, whereas supportive roles such as nurses are biased towards female figures. Our method can generate debiased results regarding the gender bias.}
    \label{fig:gender_bias_example}
\end{figure}

\begin{figure}[t!]  
    \centering
    \begin{minipage}{0.435\linewidth}  
     \centering
     \raisebox{0.01em}{
     \includegraphics[width=0.95\linewidth,trim={-2cm 2cm 0 0}]{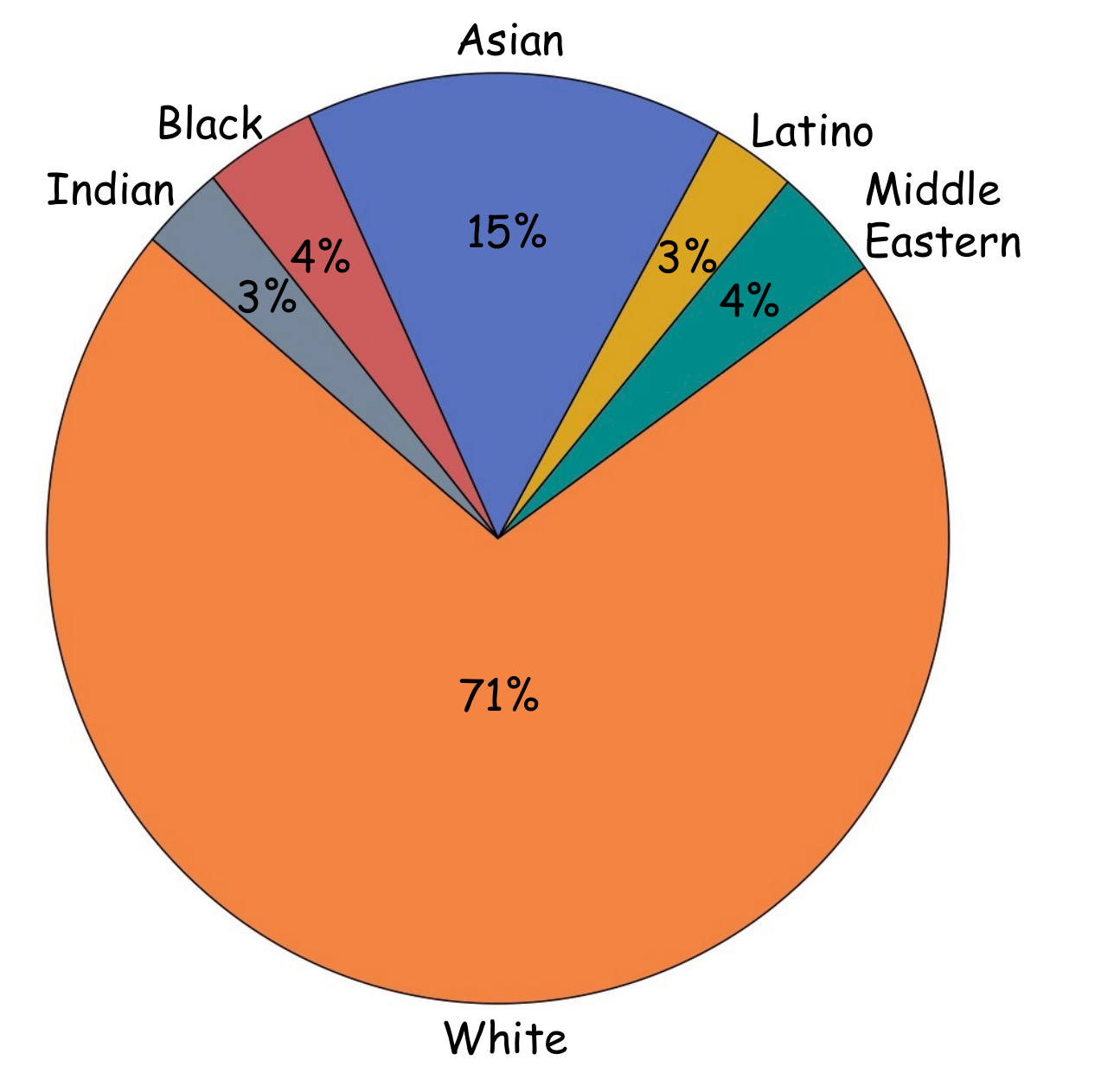}
     }
      \vspace{-0.4cm}
        \caption{Racial bias in randomly generated 1000 images with the prompt ``A photo of a worker'' using Stable Diffusion v2.1.}
        \label{fig:racial_bias}
    \end{minipage}
    \hfill
    \begin{minipage}{0.52\linewidth}
        \centering
        \raisebox{-8.25em}{ 
            \includegraphics[width=0.9\linewidth,height=0.63\linewidth]{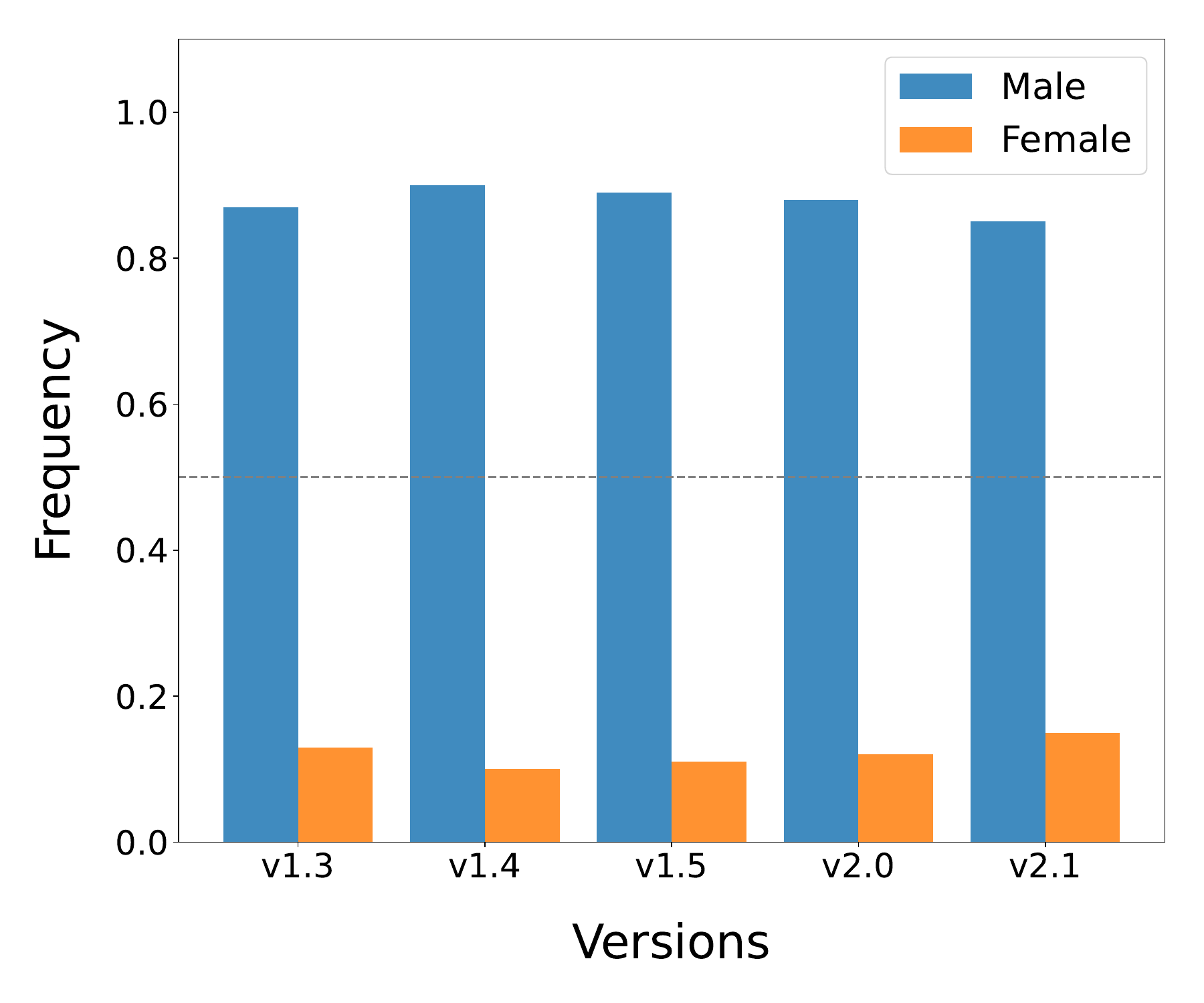}
        }
        \vspace{-0.06cm}
        \caption{Gender bias in randomly generated 1000 images with the prompt ``A photo of a CEO'' using Stable Diffusion v1.3, v1.4, v1.5, v2.0, and v2.1.}
        \label{fig:gender_bias_various_version}
    \end{minipage}
\end{figure}

In Figure \ref{fig:gender_bias_example}, we randomly generate several images of two occupations using Stable Diffusion v2.1.  Given the prompt of ``A photo of a CEO'', the generated images predominantly depict male figures, reinforcing the stereotype that leadership roles, such as CEOs and doctors, are male-dominated. On the contrary, when the prompt is ``A photo of a nurse'', the majority of generated images depict female figures, reflecting the bias that supportive roles are traditionally associated with women. Regarding racial bias, we randomly generate 1000 images using Stable Diffusion v2.1 with the prompt ``A photo of a worker''. The statistic of 1000 images depicted in Figure \ref{fig:racial_bias} shows a strong bias in racial representation, with White individuals making up 71\% of the total, while minority groups like Middle Eastern, Latino, Black, and Indian each account for only 3-4\%. This bias in DM, 
produce less accurate or fair results for underrepresented populations. We further investigate such bias situation across across different versions of DMs. We randomly generate 1000 images with the prompt of ``A photo of a CEO'' using five versions of DMs: Stable Diffusion v1.3 \citep{sd13modelcard}, v1.4 \citep{sd14modelcard}, v.1.5 \citep{sd15modelcard}, v2.0 \citep{sd20modelcard} and v2.1 \citep{sd21modelcard}. Figure \ref{fig:gender_bias_various_version} demonstrates that gender bias exists across all versions of DMs. 

This bias arises from two primary sources. First, the training data, predominantly sourced from the web, contains inherent biases. Second, bias is partially inherited from the CLIP \citep{radford2021learning} model used in the generation process. In large text-to-image DMs, these biases are even more pronounced, often manifesting in gendered associations with specific professions. Several studies have attempted to mitigate bias in DMs through retraining \citep{shen2024finetuningtexttoimagediffusionmodels, yu2020inclusive} or approaches that do not require retraining the DM~\citep{parihar2024balancingactdistributionguideddebiasing,gandikota2023unified,orgad2023editing}. In training-based methods, \citet{shen2024finetuningtexttoimagediffusionmodels} propose curating a reference dataset and aligning the distribution of generated images with this dataset. \citet{parihar2024balancingactdistributionguideddebiasing} avoid retraining the DM by leveraging demographic information embedded in the latent features of the denoising U-Net to guide generation. However, their method requires training an additional MLP-based Attribute Distribution Predictor using pseudo labels from existing attribute classifiers. This reliance on accurate attribute classifiers limits its debiasing performance. In contrast, \citet{gandikota2023unified} and \citet{orgad2023editing} employ closed-form editing to adjust concepts within DMs without retraining. While training-based methods depend on costly and quality-constrained annotated reference datasets, methods that do not require retraining the DM offer greater implementation efficiency but often exhibit lower debiasing effectiveness.
To address these limitations, we propose FairGen, a plug-and-play method that achieves debiased image generation (Figure \ref{fig:gender_bias_example}) by automatically learning attribute latent directions, mitigating the reliance on reference datasets. FairGen is composed of two components: a set of attribute adapters and a distribution indicator. Each adapter is trained to learn an attribute-specific latent direction, optimized through noise composition in a self-discovering manner, \ie, our method automatically learns attribute latent directions without relying on a labeled reference dataset. Through noise composition, our method explores and optimizes attribute directions directly from the model’s latent space, uncovering patterns without external supervision. At inference stage, the distribution indicator is then applied to select attribute-specific adapters, guiding the generative process towards the desired distribution. We comprehensively evaluate the effectiveness of our approach in debiasing gender, racial, and intersectional biases using occupational prompts. Experimental results demonstrate that our method not only achieves state-of-the-art performance in single and multiple attribute debiasing tasks but also preserves the generation quality of DM. Furthermore, we show that once FairGen is trained on one diffusion model, it can be seamlessly integrated into other models without re-tuning. Thanks to its strong transferability and plug-and-play functionality, our method offers a practical solution for both individual users and organizations, facilitating the responsible use of diffusion models in future applications.


To summarize, our main contributions are as following:
\begin{itemize}[leftmargin=2em]
    \item We propose FairGen, a novel method for debiasing DMs by learning attribute latent directions in a self-discovering manner, eliminating the reliance on the reference dataset or classifier, and thus significantly reduce the cost. 
    \item FairGen is lightweight, plug-and-play and shows good transferability across different DMs, making it more convenient to deploy in the real world.
    \item Extensive experiments show that our method achieves SOTA performance across diverse debiasing tasks while retaining
    the image generation quality.
\end{itemize}

\section{Related Work}
\vspace{-0.15cm}
\noindent\textbf{Bias in Diffusion Models.}
Diffusion models for text-to-image generation (T2I) have been observed to produce biased and stereotypical images, even when given neutral prompts. \citet{cho2022dalleval} found that Stable Diffusion (SD) tends to generate images of males when prompted with occupations, with skin tones predominantly centered around a few shades from the Monk Skin Tone Scale~\citep{monk2023monk}. \citet{seshadri2023bias} noted that SD reinforces gender-occupation biases presented in its training data. In addition to occupations, \citet{bianchi2023easily} discovered that simple prompts involving character traits and other descriptors also result in stereotypical images. \citet{luccioni2023stable} created a tool to compare generated image collections across different genders and ethnicities. Moreover, \citet{wang2023concept} introduced a text-to-image association test and found that SD tends to associate females more with family roles and males more with career-related roles.

\vspace{-0.05cm}
\noindent\textbf{Debiasing Diffusion Models by retraining.}
Before DMs, previous approaches mainly focus on debiasing GANs by assuming access to the labels of sensitive attributes and ensure no correlation exists between the decision and sensitive attributes~\citep{nam2023breaking, xu2018fairgan, van2021decaf, sattigeri2019fairness, yu2020inclusive, choi2020fair, teo2023fair, um2023fair}. More recently, regarding DMs, \citet{shen2024finetuningtexttoimagediffusionmodels} propose a distributional alignment loss to guide the characteristics of the generated images towards target distribution and use adjusted finetuning to directly optimize losses on the generated images. This method requires a reference training dataset to complete the retraining process, whereas our method does not need such reference dataset, which largely reduces annotation costs. \citet{yu2020inclusive} enhance fairness in text-to-image synthesis by incorporating reference images. and leverage visual exemplars to more effectively represent attributes that are difficult to describe in words, such as nuanced variations in skin tones. \citet{li2024selfdiscoveringinterpretablediffusionlatent} find a interpretable latent vector for a given concept, and achieve debiasing by manipulating the h-space inside DMs during inference stage.

\vspace{-0.05cm}
\noindent\textbf{Debiasing Diffusion Models without retraining.}
\citet{parihar2024balancingactdistributionguideddebiasing} propose Distribution Guidance (DG), which guides the generated images to follow the prescribed attribute distribution. Although DG does not require retraining DMs, it requires training an Attribute Distribution Predictor (ADP) that maps the latent features to the distribution of attributes. Since ADP is trained with pseudo labels generated from existing attribute classifiers, the performance of DG is largely constrained by the accuracy of attribute classifiers. \citet{gandikota2023unified} and \citet{orgad2023editing} use closed-form editing approach to edits concepts inside DM without training. Despite the efficiency, these methods are less effective than training-based approachs.

\vspace{-0.05cm}
\noindent\textbf{Controling concepts in Diffusion Models.}
Our work is inspired by recent work on controling concepts~\citep{gandikota2023erasingconceptsdiffusionmodels,gandikota2023conceptslidersloraadaptors} in diffusion models. \citet{gandikota2023erasingconceptsdiffusionmodels} propose to fine-tune the DM to erase an undesired concept while maintaining the model’s behavior and capabilities, such as removing the NSFW (not safe for work) contents.
ConceptSlider~\cite{gandikota2023conceptslidersloraadaptors} also fine-tunes a LoRA adapter in a diffusion model to control the concept in the generated images.
Unlike ESD that focuses on erasing concepts and ConceptSlider aims at precisely adjusting the extent of a concept (\eg, controling the size of eyes), this work aims to debias the Diffusion Models, \ie., controling the portion of generated images for different attributes such as male and female.
Additionally, ConceptSlider learns an adapter per concept and use a scaling factor to control the extent of a concept (\eg, size of eye) while we learn an adapter per direction (\eg., female or male) in an attribute/concept (\eg, gender) and design an inference pipeline with an indicator to automatically select adapters based on a given distribution.
\section{Preliminary}
\vspace{-0.15cm}
\textbf{Latent Diffusion Models (LDMs)} perform the diffusion process within the
latent space~\citep{rombach2022high}.
During training, noise is added to the encoded latent representation of the input image \( x \), resulting in a noisy latent code \( z_t \) at each time step \( t \). In the pretraining stage, an autoencoder framework is employed to map images into a lower-dimensional latent space via an encoder: \( z = \mathcal{E}(x) \). The decoder then reconstructs images from these latent codes: \( x \approx \mathcal{D}(\mathcal{E}(x)) \). This process ensures that the latent space retains the essential semantic information of the image. The training objective of the diffusion model in the latent space is given by:
\begin{equation}
\mathcal{L}_{\mathrm{LDM}} = \mathbb{E}_{z \sim \mathcal{E}(x), c, \epsilon \sim \mathcal{N}(0,1), t}\left[\left\|\epsilon - \epsilon_{\theta}\left(z_t, c,t\right)\right\|_2^2\right],
\end{equation}
where \( \epsilon \) is Gaussian noise sampled from a normal distribution \( \mathcal{N}(0,1) \), \( \epsilon_{\theta} \) is the denoising network, and \( c \) represents any conditioning embeddings (\eg, text or class labels). At the inference stage, a latent code \( z_T \) is sampled from Gaussian noise at the initial timestep \( T \). The denoising network \( \epsilon_{\theta} \) is then applied iteratively to remove the noise over several steps, generating a denoised latent representation \( z_0 \). Finally, the pretrained decoder reconstructs the image from the denoised latent code: \( \hat{x}_0 \approx \mathcal{D}(z_0) \), where \( \hat{x}_0 \) is the generated output image.

\noindent\textbf{Classifier-free Guidance} aims to modulate image generation by steering the probability distribution towards data that is more probable according to an implicit classifier \( p(c \mid z_t) \)~\citep{ho2022classifierfree}. It operates at the inference phase and the model is jointly trained on both conditional and unconditional denoising tasks. During inference, both the conditional and unconditional denoising scores are derived from the model. The final score \( \Tilde{\epsilon}_\theta(z_t, c, t) \) is then adjusted by weighting the conditioned score more heavily relative to the unconditioned score using a guidance scale \( \alpha > 1 \).
\begin{align}
    \Tilde{\epsilon}_\theta(z_t, c, t) = \epsilon_\theta(z_t,t) + \alpha(\epsilon_\theta(z_t, c, t) - \epsilon_\theta(z_t,t))
\end{align}
The inference process begins with sampling a latent variable \( z_T \sim \mathcal{N}(0,1) \), which is subsequently denoised using \( \Tilde{\epsilon}_\theta(z_t, c, t) \) to obtain \( z_{t-1} \). The denoising is performed iteratively until obtaining \( z_0 \). Finally, the decoder transforms the latent representation \( z_0 \) back into image space: \( x_0 \gets \mathcal{D}(z_0) \).

\section{Method}


\begin{figure*} [tb!]
   \begin{center}
   \vspace{-0.2cm}
   \includegraphics[width=0.95\linewidth]{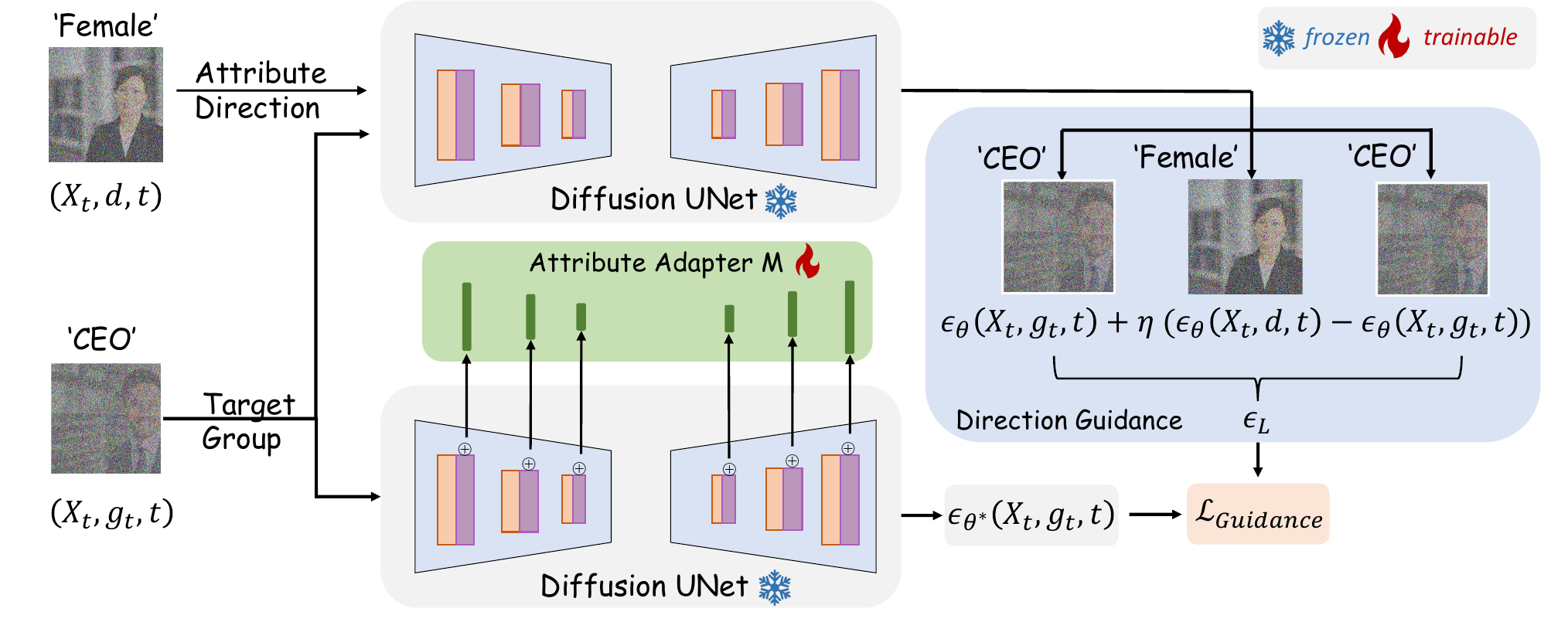} 
   \end{center}
   \vspace{-0.6cm}
    \caption{Overview of FairGen's training pipeline. Attribute-specific adapters ($M$) are attached to the cross-attention layers in the denoising UNet. 
    Target group and attribution direction are fed into the DM for composing the noise predictions (Eq. \ref{essence}), which is used as self-discovering attribute direction guidance to optimize the adapters.
    }
   \label{fig: train pipelihne} 
   \end{figure*}

\begin{figure*} [tb!]
   \begin{center}
   \includegraphics[width=0.95\linewidth]{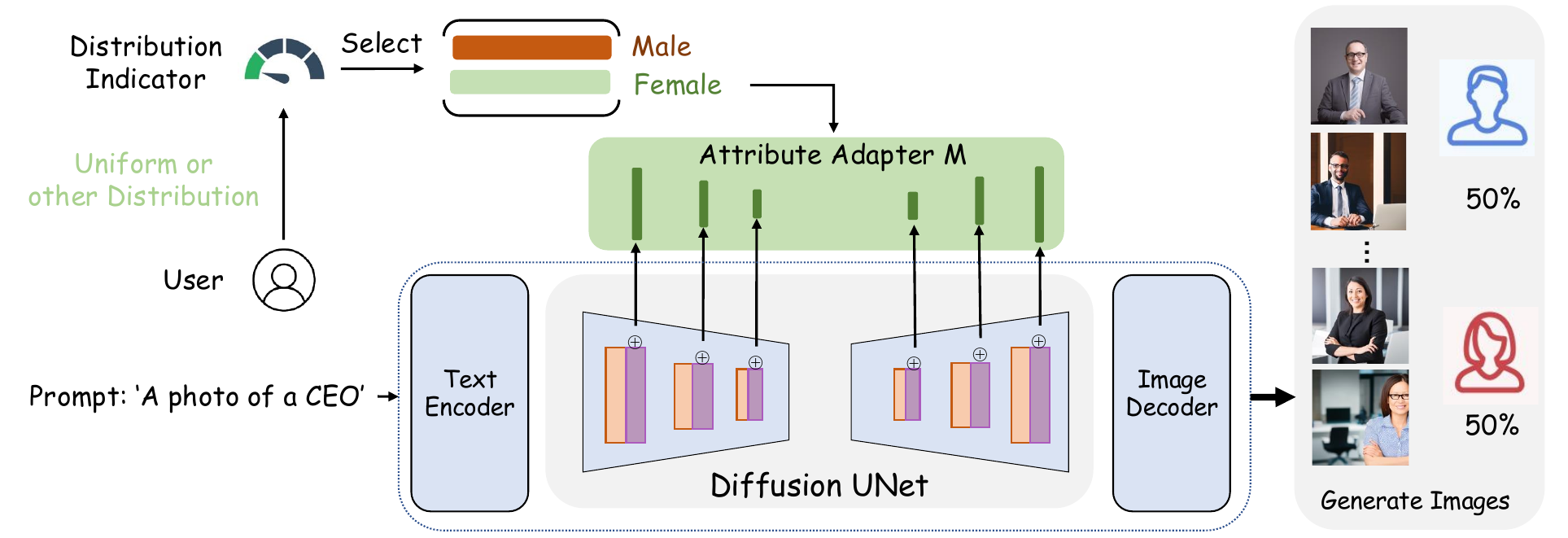} 
   \end{center}
   \vspace{-0.6cm}
    \caption{Overview of FairGen's inference pipeline. The distribution indicator is generated according the prescribed distribution. Then, it is multiplied by the set of attribute adapters matrices to select attribute matrix adapter. The selected adapter is integrated to the DM with no overhead, guiding the generation towards the prescribed distribution.}
   \label{fig:inferencepipelihne} 
   \end{figure*}

Inspired by recent work that fine-tune diffusion models for erasing concepts~\citep{gandikota2023erasingconceptsdiffusionmodels} and controlling the level of concepts~\cite{gandikota2023conceptslidersloraadaptors} (\eg, ages, size of eyes), we aim at reducing the bias in the DM by attaching and learning a set of light-weight adapters, each of which represents a category of an attribute (\eg, female of gender), guiding the DM towards an attribute latent direction. Unlike previous work that relies on additional reference datasets and has to finetune the whole DM~\citep{shen2024finetuningtexttoimagediffusionmodels}, we instead optimize the attached adapters via noise composition through a self-discovering process (detailed in Section \ref{sec:selfdiscover}). This significantly reduces the cost in computation and data. During the inference stage (Section \ref{sec:infer}), given a predefined target distribution (\eg, uniform), we introduce a distribution indicator implemented by a gating function to select one corresponding adapter which will be attached to the DM for generating image. In this way, the set of generated images will follow the predefined target distribution, and thus are not biased to some categories of an attribute (if the predefined target distribution is uniform). Our training and inference diagrams are illustrated in Figure \ref{fig: train pipelihne} and Figure \ref{fig:inferencepipelihne}. In the following sections, we start elaborating our method in single attribute settings and then extend it to more general ones.

\subsection{Attaching Light-weight attribute-specific Adapters}\label{sec:adapters}

To debias a DM for generating images of a given attribute that contains several categories (\eg, male and female are two categories of gender attribute), we first aim at equip the DM with skills of generating images for each category. To achieve this, we attach a light-weight adapter per category in each layer of the DM inspired by parameter-efficient fine-tuning (PEFT) instead of finetuning the whole model to acheive a good trade-off between performance and computational cost. In this work, we use the 1-dim adapter \citep{lyu2024onedimensionaladapterruleall} and only add the adapter to each cross-attention layers of the denoising U-Net, as shown in Figure \ref{fig: train pipelihne} as we find that attaching adapters to all layers does not help and will also increase the computational cost.

Specifically, for the $i$-th cross-attention layer parameterized by $\boldsymbol{W}_i \in \mathbb{R}^{m\times n}$ in the denoising U-Net, we attach an adapter to the layer to guide the attribute towards a certain category. The adapter consists of two 1-dim vectors: $\boldsymbol{p} \in \mathbb{R}^{m}$ and $\boldsymbol{q} \in \mathbb{R}^{n}$. The forward process of the $i$-th cross-attention layer would be updated from $\boldsymbol{y}_i = \boldsymbol{W}_i \boldsymbol{x}_i$ to
$\boldsymbol{y}_i = \boldsymbol{W}_i \boldsymbol{x}_i + (\boldsymbol{q}_i^T \boldsymbol{x}_i) \cdot \boldsymbol{p}_i$.
$\boldsymbol{x}_i \in \mathbb{R}^n$ and $\boldsymbol{y}_i \in \mathbb{R}^m$ represent the input and output of the layer, and superscript $T$ indicates transposition. Thus, adapters in all ($r$) cross-attention layers for an attribute category $d$ are:
\begin{equation}
M_d = \boldsymbol{Q}^T \boldsymbol{P},
\end{equation}
where 
$\boldsymbol{Q} = \left[\boldsymbol{q}_1, \boldsymbol{q}_2, \ldots, \boldsymbol{q}_r\right], \boldsymbol{P} = \left[\boldsymbol{p}_1, \boldsymbol{p}_2, \ldots, \boldsymbol{p}_r\right].$
Each column vector in $\boldsymbol{Q}$ and $\boldsymbol{P}$, we pad 0 to their end if their dimensions are not the same.

\subsection{Optimizing Adapters via self-discovering process}\label{sec:selfdiscover}
One straightforward way to optimize the attached adapters is to collect a unbiased reference dataset as in prior work \citep{shen2024finetuningtexttoimagediffusionmodels}. However, it is expensive to collect such dataset and the quality of the dataset would also limites the performance. To this end, we propose to train the adapters in a self-discovering manner. 


Given a target group $g_t$ (for example, ``CEO'') and model $\theta$, we want to optimize the adapters such that the model attached with the adapters generate image $X$ towards certain attribute category $d$ (for example, ``male'' or ``female'' ) when conditioned on $g_t$:
\begin{align}
\label{eq:7}
    P_{\theta^*}(X | g_t) \gets P_{\theta}(X | g_t) \left({ P_{\theta}(d | X)}\right)^{\eta},
\end{align}
where $P_{\theta}(X | g_t)$ represents the distribution generated by the original model when conditioned on $g_t$, and $\theta^*$ represents the new model equipped with the adapters of $d$. Note that $g_t$ can be set to an empty string ``'' so that the model will be debiased for all possible groups.

Applying the Bayes Formula, $P(d|X)=\frac{P(X|d)P(d)}{P(X)}$ to Eq. \ref{eq:7}, taking logarithm on both sides, we are able to derive that 
the gradient of the log probability $\nabla \log P_{\theta^*}(X|g_t)$ would be proportional to:
\begin{align}
    \label{essence}
    \nabla \log P_{\theta}(X | g_t) + \eta \left(\nabla \log P_{\theta}(X|d) - \nabla \log P_{\theta}(X|g_t)\right)
\end{align}
Based on Tweedie's formula \citep{tweedie} and the reparametrization trick of Classifier-free guidance \citep{ho2022classifierfree}, we introduce a time-varying noising process and represent each score (gradient of log probability) as a denoising prediction $\epsilon(X_t,c_t,t)$, which leads to our learning objective for adapters of category $d$ of an attribute:
\begin{align}
\begin{split}
    \epsilon_{\theta^*}(X_t, g_t, t)  \gets \epsilon_{\theta}(X_t, g_t, t) +
    \eta\left(\epsilon_{\theta}(X_t, d, t) - \epsilon_{\theta}(X_t, g_t, t) \right).
\end{split}
\end{align}
Therefore, the guidance loss for optimizing adapters of category $d$ can be defined as follows:
\begin{align}\label{eq:guidance}
\begin{split}
    \mathcal{L}_{\text {Guidance }}  =\mathbb{L}_{x_t, t}\left[\left\|\epsilon_{\theta^*}(X_t, g_t, t)-\epsilon_L\right\|^2\right],\\
    \epsilon_L=\epsilon_{\theta}(X_t, g_t, t) \; + 
     \eta\left(\epsilon_{\theta}(X_t, d, t) - \epsilon_{\theta}(X_t, g_t, t) \right),
\end{split}
\end{align}
where the goal is to align the noise $\epsilon_{\theta^*}(X_t, g_t, t)$ of the $\theta^*$ (DM with adapters of category $d$) and the noise composition $\epsilon_L$ of group $g_t$ and category $d$ from the frozen DM $\theta$. And hence, our adapters optimization does not require any additional data.

\subsection{Inference with distribution indicator.}\label{sec:infer}
After the optimization, we obtain a set of adapters $\mathcal{M} = \{\boldsymbol{M}_1, \boldsymbol{M}_2, \ldots, \boldsymbol{M}_t\}$ for $t$ categories of a given attribute. For instance, $t=2$ for gender bias and $t=4$ for racial bias in our case. As shown in Figure \ref{fig:inferencepipelihne}, at the inference stage, we introduce a distribution indicator $\boldsymbol{h}$. Given a prescribed distribution ${f^a_{\theta}}$ (\eg, uniform distribution), we define its Probabilistic Mass Function (PMF) as follows:
\begin{equation}
    P(X = x) = f^a_{\theta}(x) = 
\begin{cases}
\frac{1}{t} & \text{for } x \in \{1, 2, \dots, t\}, \\
0 & \text{otherwise},
\end{cases}
\end{equation}
where $t$ is the number of possible categories for each attribute, and $\{1, 2, \dots, t\}$ represents the set of all possible values for the random variable $X$. The parameter $\theta$ controls the shape of the prescribed distribution, and in the case of the uniform distribution, $\theta$ implies that all outcomes in the set have equal probability, \ie, $\frac{1}{t}$.

Then, we randomly sample an index $k$ from the prescribed distribution ${f^a_{\theta}}$. The distribution indicator $\boldsymbol{h} \in \mathbb{R}^{t}$ is formulated to reflect the chosen index $k$ as follows:

\begin{equation}
\boldsymbol{h}_i := \begin{cases} 
1 & \text{if } i = k, \\
0 & \text{if } i \neq k,
\end{cases}
\end{equation}

where $i \in \{1, 2, \dots, t\}$, and $k$ is the sampled index based on the distribution ${f^a_{\theta}}$. The indicator $\boldsymbol{h}$ is a one-hot vector, where the $k$-th element is 1, indicating the sampled outcome, and all other elements are 0. After obtaining the distribution indicator, it is multiplied by the set of trained attribute matrix adapters. Subsequently, the final weight change $\Delta \boldsymbol{W}$ is given by:
\begin{equation}
    \Delta \boldsymbol{W} = \boldsymbol{h} \cdot \mathcal{M},
\end{equation}
and the model is updated as $\boldsymbol{W} \gets \boldsymbol{W} + \alpha \Delta \boldsymbol{W}$, where $\alpha$ is a scaling factor controlling the strength of the guidance.

\subsection{Debiasing Multiple Attributes (intersectional debiasing)}



\begin{figure*}[t!]
   \begin{center}
   \includegraphics[width=0.92\linewidth]{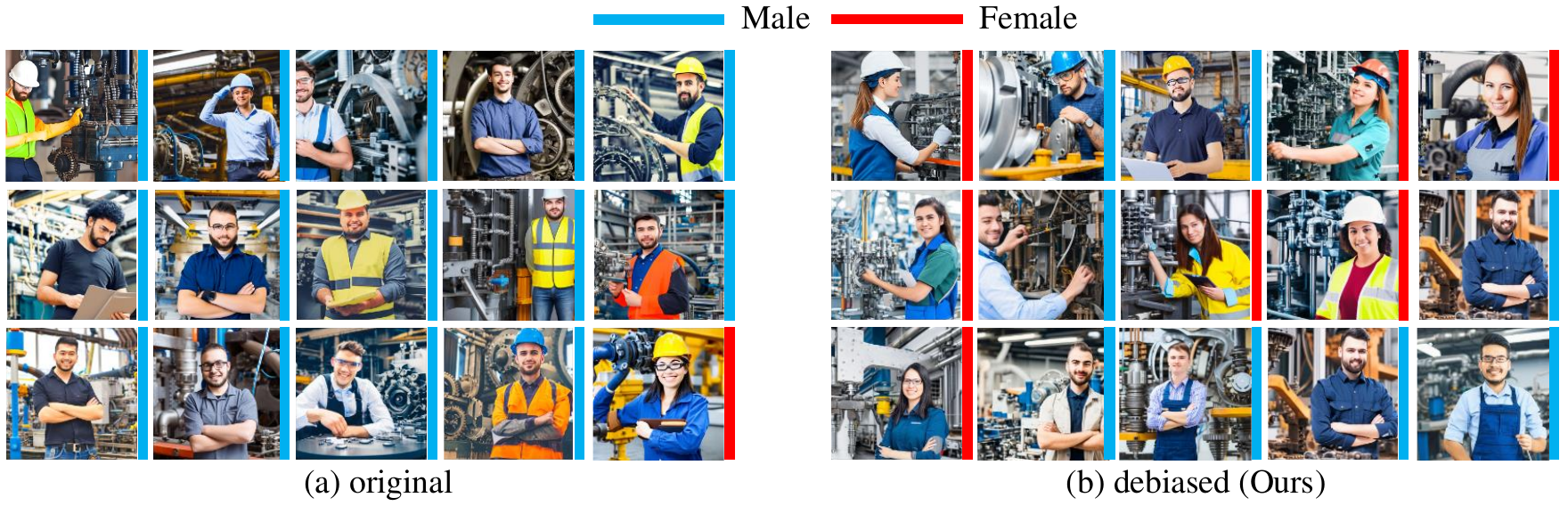} 
   \end{center}
   \vspace{-0.7cm} 
      \caption{Images generated from the original Stable Diffusion model (left) and FairGen (right) for gender debiasing with the prompt ``A photo of a worker''. The gender distribution shifts from a male-to-female ratio of 13:2 in the original model to 8:7 with FairGen.}
\label{fig:genderdebiasvisual} 
   \end{figure*}

Our method can be inherently extended to debiasing multiple attributes in diffusion models (DM). Specifically, in the case of multiple attribute debiasing, the adapter for each attribute should not interfere with the others. Otherwise, the most recently trained adapter could degrade the performance of previously learned adapters. For each attribute to be debiased, we denote a set of adapter parameters as $\left\{\boldsymbol{P}_t, \boldsymbol{Q}_t\right\}$. We have 
$\boldsymbol{P}_t = \left[\boldsymbol{p}_t^1, \boldsymbol{p}_t^2, \ldots, \boldsymbol{p}_t^r\right], 
\boldsymbol{Q}_t = \left[\boldsymbol{q}_t^1, \boldsymbol{q}_t^2, \ldots, \boldsymbol{q}_t^r\right]$.

To avoid interference between attribute adapters, we extend Eq. \ref{eq:guidance} by introducing an orthogonal regularization loss that regularizes the vector subspace spanned by each $\boldsymbol{P}_t$ and $\boldsymbol{Q}_t$ to be orthogonal to each other:
\vspace{-0.2cm}
\begin{equation}\label{eq:or}
    \mathcal{L}_{\text{orth}} = \sum_{i=1}^{t-1} \left( \boldsymbol{P}_i \times \boldsymbol{P}_t + \boldsymbol{Q}_i \times \boldsymbol{Q}_t \right),\\
\end{equation}
\vspace{-0.2cm}
\begin{equation}
    \mathcal{L} = \mathcal{L}_{\text{Guidance}} + \gamma \cdot \mathcal{L}_{\text{orth}}.
\end{equation}
\vspace{-0.2cm}


\section{Experiment}
In this section, we first describe the experimental details and evaluation metrics, then present a quantitative and qualitative analysis of our method. Please refer to Appendix for more results. 

\vspace{-0.1cm}
\subsection{Experimental Details}
\vspace{-0.1cm}
\paragraph{Implementation details.}
We use Stable Diffusion v2.1 for all methods. We employ the prompt template ``\textit{a photo of the face of a} \{occupation\}\textit{, a person}''. At inference time, for each bias, we generate 100 images per occupation across 100 occupations, resulting in a total of 10,000 images. We set $\eta = \alpha = 0.3$,  and train for 1000 iterations with a learning rate of 1e-5. For gender bias, we use the CelebA \citep{liu2015faceattributes} dataset to train a binary classifier with two categories:\{male,female\}. For racial bias, we use the FairFace \citep{fairface-model} dataset to train a classifier with the following four categories: \textcolor{red1}{WMELH}=\{White, Middle Eastern, Latino Hispanic\}, \textcolor{blue1}{Asian}=\{East Asian, Southeast Asian\}, Black, and \textcolor{brown}{Indian}. Please refer to the supplementary for more details.



\vspace{-0.4cm}
\paragraph{Compared methods.}
In this work, we compare our method with recent state-of-the-art (SOTA) methods, including two retraining-based methods, \textbf{Finetuning for Fairness (F4Fair)}~\citep{shen2024finetuningtexttoimagediffusionmodels} and \textbf{Inclusive Text-to-Image Generation (ITI-GEN)}~\citep{zhang2023itigeninclusivetexttoimagegeneration}, an approach that does not require retraining DM, \textbf{H-Distribution Guidance (H Guidance)}~\citep{parihar2024balancingactdistributionguideddebiasing}, a closed-form editing approach, \textbf{Unified Concept Editing (UCE)}~\citep{gandikota2023unified} and an h-space manipulation approach, \textbf{Interpretable Diffusion}~\citep{li2024selfdiscoveringinterpretablediffusionlatent}. Please refer to Appendix for more detailed introduction of these methods. 

\vspace{-0.1cm}
\subsection{Evaluation Metrics}
\vspace{-0.1cm}
We evaluate all methods in three metrics: 

\vspace{0.05cm}
\noindent\textbf{Fairness Discrepancy (FD).} Following prior work\citep{parihar2024balancingactdistributionguideddebiasing}, we adopt the {Fairness Discrepancy} (FD) metric. For an attribute $a$ and target distribution ${p^a_{\theta}}$, we use a high-accuracy classifier $\mathcal{C}_a$ to compute the fairness performance:
$||{p^a_{\theta}} - \mathbb{E}_{\mathbf{x} \sim p_\theta(\mathbf{x})}(\mathbf{y})||_2$
where $\mathbf{y}$ is the softmax output of $\mathcal{C}_a(\mathbf{x})$. The target distribution ${p^a_{\theta}}$ can be any user-defined vector, typically uniform. A lower FD score indicates a closer match to the target distribution.

\vspace{0.05cm}
\noindent\textbf{CLIP\textsubscript{sim}.} Besides fairness, the debiased model should maintain the ability to generate images that are semantically close to their text prompts. Therefore, following \citep{shen2024finetuningtexttoimagediffusionmodels}, we report the CLIP similarity score CLIP\textsubscript{sim} between the generated image and its prompt.

\vspace{0.05cm}
\noindent\textbf{BRISQUE.} The generated image quality is also important as the debiasing process should not influence the image generation ability. We use the BRISQUE metric for evaluate the quality of the generated images.

\begin{figure*} [tb!]
   \begin{center}
   \includegraphics[width=0.93\linewidth]{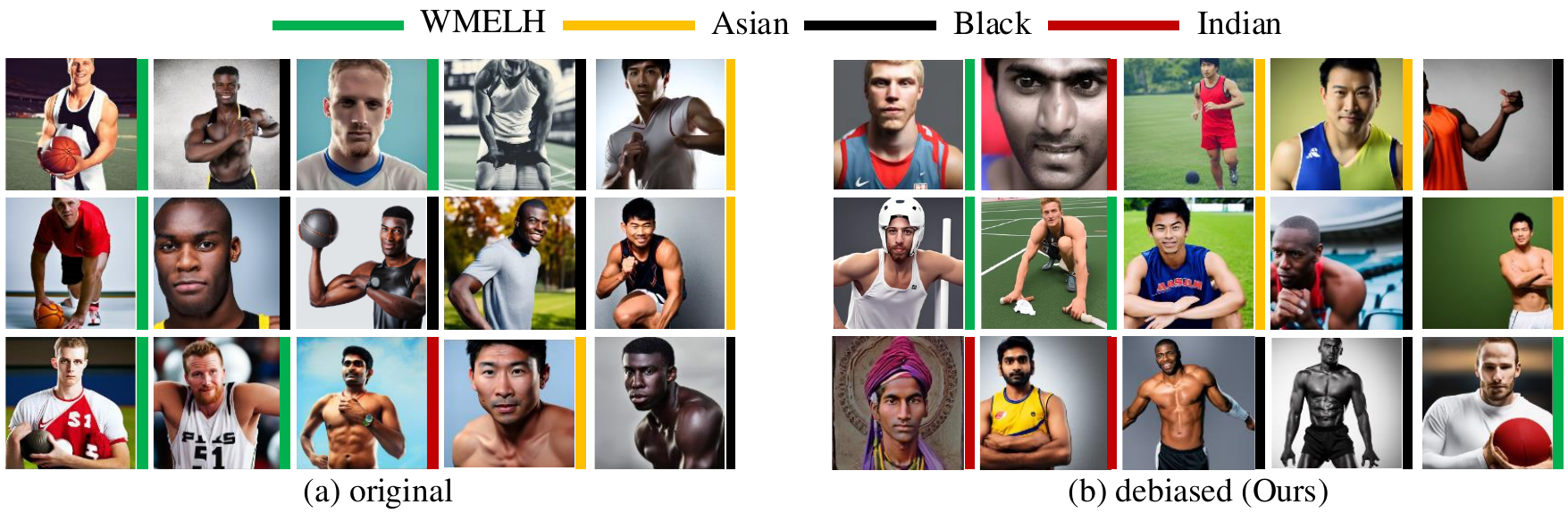} 
   \end{center}
   \vspace{-0.7cm} 
    \caption{Images generated from the original Stable Diffusion model (left) and FairGen (right) for racial debiasing with the prompt \textit{``A photo of a sportsman"}. The racial group distribution shifts from \textcolor{red1}{WMELH} : \textcolor{blue1}{Asian} : Black : \textcolor{brown}{Indian} = 7:2:5:1 in the original model to 4:4:4:3 with FairGen.}
   \label{fig: racial bias visual} 
   \vspace{-0.5cm} 
   \end{figure*}

\subsection{Results}

\vspace{-0.1cm}
\paragraph{Comparisons in gender debiasing.}
Table \ref{tab:gender_bias_results} demonstrates that our method outperforms others in mitigating gender bias over two predefined distributions: ${f^1_{\theta}} = (0.5, 0.5)$, representing equal likelihood of male and female, and ${f^2_{\theta}} = (0.2, 0.8)$, where male and female have a 20\% and 80\% probability, respectively. Original SD model exhibits high FD scores (0.424 at $f^1_{\theta}$ and 0.847 at $f^2_{\theta}$), indicating significant bias towards one gender. 
While previous methods like F4Fair, ITI-GEN, H Guidance, and UCE reduce bias to some extent, our method achieves the most reduction, with FD scores of 0.003 at $f^1_{\theta}$ and 0.005 at $f^2_{\theta}$. Additionally, our method preserves semantic similarity (CLIP\textsubscript{sim} of 0.38) and image quality (BRISQUE scores of 38.46 and 38.72), matching or slightly surpassing the Original SD model. The qualitative results in Figure \ref{fig:genderdebiasvisual} further demonstrate that our approach effectively mitigates gender bias without compromising image quality or semantic coherence.

\begin{table}[t]
\caption{
    Comparisons of our method to the SOTA methods in gender bias over two predefined distributions: ${f^1_{\theta}} = (0.5, 0.5)$ and ${f^2_{\theta}} = (0.2, 0.8)$, representing the probability of male and female respectively.
    }
    \vspace{-0.3cm}
    \label{tab:gender_bias_results}
    \resizebox{0.44\textwidth}{!}
    {
    \begin{tabular}{l|cc|cc|cc}
    \toprule
    \multirow{2}{*}{Method} & \multicolumn{2}{c|}{FD $\downarrow$} & \multicolumn{2}{c|}{CLIP\textsubscript{sim} $\uparrow$} & \multicolumn{2}{c}{BRISQUE $\uparrow$} \\
    \cmidrule{2-7}
     & $f^1_{\theta}$ & $f^2_{\theta}$ & $f^1_{\theta}$ & $f^2_{\theta}$ & $f^1_{\theta}$ & $f^2_{\theta}$ \\
    \midrule
    Original SD  & 0.424  & 0.847  & \textbf{0.38}  & \textbf{0.38}  & \textbf{38.65}  & 38.69  \\
    \midrule
    F4Fair       & 0.165  & 0.387  & 0.36  & 0.35  & 38.21  & 37.64  \\
    ITI-GEN       & 0.093  & 0.472  & 0.32  & 0.33  & 37.62  &   37.96\\
    H Guidance   & 0.118  & 0.398  & 0.31  & 0.32  & 38.54  & 38.65  \\
    UCE          & 0.284  & 0.536  & 0.36  & 0.29  & 37.12  & 36.54  \\
    InterDiff    & 0.177 &  0.386 & 0.35  &  0.34 &  38.32 &  38.04\\
    \rowcolor{lightblue}  
    Ours         & \textbf{0.003} & \textbf{0.005} & \textbf{0.38} & \textbf{0.38} & 38.46 & \textbf{38.72} \\
    \bottomrule
    \end{tabular}
    }
\end{table}

\vspace{-0.5cm}
\paragraph{Comparisons in racial debiasing.}
Table \ref{tab:racial_bias_results} compares methods over two distributions: ${f^1_{\theta}} = (0.25, 0.25, 0.25, 0.25)$, representing equal probabilities for \textcolor{red1}{WMELH}, \textcolor{blue1}{Asian}, Black, and \textcolor{brown}{Indian}, and ${f^2_{\theta}} = (0.4, 0.3, 0.2, 0.1)$, with higher probabilities for \textcolor{red1}{WMELH} and \textcolor{blue1}{Asian}, and lower probabilities for Black and \textcolor{brown}{Indian}. Original SD model shows significant racial bias, with FD scores of 0.384 for ${f^1_{\theta}}$ and 0.497 for ${f^2_{\theta}}$, indicating poor calibration to either distribution.

\begin{table}[t]
\caption{Comparisons of our method to the SOTA methods in racial bias over two distributions: ${f^1_{\theta}} = (0.25, 0.25, 0.25, 0.25)$ and ${f^2_{\theta}} = (0.4, 0.3, 0.2, 0.1)$, representing the probability of \textcolor{red1}{WMELH}, \textcolor{blue1}{Asian}, Black, and \textcolor{brown}{Indian} respectively.}
    \vspace{-0.3cm}
    \label{tab:racial_bias_results}
    \resizebox{0.44\textwidth}{!}
    {
    \begin{tabular}{l|cc|cc|cc}
    \toprule
    \multirow{2}{*}{Method} & \multicolumn{2}{c|}{FD $\downarrow$} & \multicolumn{2}{c}{CLIP\textsubscript{sim} $\uparrow$} & \multicolumn{2}{c}{BRISQUE $\uparrow$} \\
    \cmidrule{2-7}
                            &  \( f^1_{\theta} \) &  \( f^2_{\theta} \) & \( f^1_{\theta} \) & \( f^2_{\theta} \) & \( f^1_{\theta} \) & \( f^2_{\theta} \) \\
    \midrule
    Original SD             &   0.384    &    0.497      &  \textbf{0.46}      &   \textbf{0.41}    &      38.94    &   38.65      \\
    \cmidrule{1-7} 
    F4Fair                  &   0.220    &    0.305      &  0.43      &   0.40    &      38.80    &   38.45      \\
    ITI-GEN       & 0.131  & 0.542  & 0.31  & 0.31  & 36.61  &   37.09\\
    H Guidance              &   0.150    &    0.285      &  0.41      &   0.37    &      38.65    &   38.60      \\
    UCE                     &   0.290    &    0.460      &  0.45      &   0.39    &      37.90    &   37.80      \\
    InterDiff                     &   0.150    &    0.391      &  0.40      &   0.38    &      37.64    &   37.82      \\
    \rowcolor{lightblue}
    Ours                    &   \textbf{0.095}    &    \textbf{0.150}      &  \textbf{0.46}      &   \textbf{0.41}    &      \textbf{38.96}    &   \textbf{38.66}      \\
    \bottomrule
    \end{tabular}
    }
\end{table}

Debiasing methods such as F4Fair, H Guidance, and UCE reduce this bias, with H Guidance achieving relatively lower FD scores. However, our method performs the best, reducing bias to 0.095 for ${f^1_{\theta}}$ and 0.150 for ${f^2_{\theta}}$. In terms of semantic similarity, the Original SD model sets a strong baseline (0.46 for ${f^1_{\theta}}$ and 0.41 for ${f^2_{\theta}}$), and our method maintains this high alignment, ensuring that debiasing does not impair semantic accuracy. Regarding image quality, our approach slightly improves upon the Original SD model, achieving the highest scores (39.10 for ${f^1_{\theta}}$ and 38.90 for ${f^2_{\theta}}$). Again, the qualitative results in  Figure \ref{fig: racial bias visual} further verify that our method outperforms others in reducing racial bias while preserving both semantic similarity and image quality.

\vspace{-0.5cm}
\paragraph{Comparisons in intersectional debiasing}


\begin{table*}[ht!]
\centering
\caption{Transferability of proposed method across different DMs}
\vspace{-0.3cm}
\label{tab:transfer}
\resizebox{0.6\textwidth}{!}
    {
\begin{tabular}{l|l|ccc|ccc}
\toprule
\multirow{2}{*}{Train} & \multirow{2}{*}{Test} & \multicolumn{3}{c|}{Gender} & \multicolumn{3}{c}{Racial} \\
\cmidrule{3-8}
& & FD $\downarrow$ & CLIP\textsubscript{sim} $\uparrow$ & BRISQUE $\uparrow$ & FD $\downarrow$ & CLIP\textsubscript{sim} $\uparrow$ & BRISQUE $\uparrow$ \\
\midrule
v2.1 & v1.4 & 0.006 & 0.37 & 38.40 & 0.097 & 0.45 & 38.90 \\
v2.1 & v1.5 & 0.006 & 0.37 & 38.42 & 0.096 & 0.45 & 38.92 \\
v2.1 & v2.0 & 0.008 & 0.38 & 38.45 & 0.095 & 0.46 & 38.95 \\
\rowcolor{lightblue}
v2.1 & v2.1 & \textbf{0.003} & \textbf{0.38} & \textbf{38.46} & \textbf{0.095} & \textbf{0.46} & \textbf{38.96} \\
\bottomrule
\end{tabular}
}
\end{table*}

We also consider a more complex challenge of intersectional debiasing, \ie, jointly debiasing both gender and racial biases. The target distribution $f_\theta$ is set to be uniform for both gender and racial bias. Table \ref{tab:intersectional_bias_results} shows that Original SD model exhibits significant bias with an FD score of 0.214. While F4Fair and H Guidance reduce this bias to 0.145 and 0.130, respectively, our method achieves a much lower FD score of 0.047, reflecting a substantial improvement in fairness.

\begin{table}[t!]
\centering
\caption{Evaluation of mitigating intersectional bias.}
\vspace{-0.3cm}
\label{tab:intersectional_bias_results}
\resizebox{0.38\textwidth}{!}
    {
\begin{tabular}{l|c|c|c}
\toprule
{Method} & { FD $\downarrow$} & {CLIP\textsubscript{sim}$\uparrow$} & { BRISQUE $\uparrow$} \\
\midrule
Original SD             &  0.214     &     0.35     &   39.24         \\
\cmidrule{1-4} 
F4Fair                  &  0.145     &     0.36     &   38.90         \\
ITI-GEN                  &  0.098     &     0.32     &   38.01         \\
H Guidance              &  0.130     &     0.33     &   38.75         \\
UCE                     &  0.180     &     0.34     &   38.60         \\
InterDiff                     &  0.158     &     0.35     &   38.67         \\
\rowcolor{lightblue}
Ours                    &  \textbf{0.047}     &     \textbf{0.36}     &   \textbf{39.50}         \\
\bottomrule
\end{tabular}
}
\end{table}

For semantic similarity, the Original SD scores 0.35, and F4Fair slightly improves it to 0.36. Our method matches this performance, maintaining semantic coherence while reducing bias. Regarding image quality, the Original SD has a BRISQUE score of 39.24, while our method improves it to 39.50, indicating enhanced perceptual quality.
These results demonstrate that our method excels at jointly debiasing both gender and racial biases, significantly reducing bias without sacrificing semantic accuracy or image quality. This highlights its robustness and practicality in real-world settings where multiple biases are present.

\vspace{-0.5cm}
\paragraph{Transferability across different DMs.}
We further evaluate the transferability of our method by training it with Stable Diffusion v2.1 and testing it over other versions, and results are reported in Table \ref{tab:transfer}.
From the table we can see that the performance of in all metrics are close to the optimal results achieved when training and testing are performed using the same model version (v2.1). While there is a slight increase in FD when tested on v1.4, v1.5, and v2.0, the differences are minor, indicating that our method can effectively generalize across different model versions. This robustness highlights the method's capability to transfer learned features across varying model conditions.

\subsection{Ablation Study}
\vspace{-0.1cm}
\paragraph{Attaching adapters in U-Net.}
The results in Table \ref{tab:attach_ablation} illustrate the impact of attaching adapters to different layers of the U-Net in the DM. When adapters are attached to all layers, the fairness score (FD) is 0.165, with a CLIP similarity score of 0.33 and a BRISQUE score of 35.00. The best results are obtained when adapters are attached only to the CA layers. This configuration significantly reduces the FD score to 0.047, increases semantic similarity (CLIP\textsubscript{sim} = 0.36), and enhances image quality with a BRISQUE score of 39.50. These highlights that focusing the adapters on the cross-attention layers leads to the most substantial improvements in both fairness and image quality.

\begin{table}[t]
    \centering
\caption{Ablation of layers that are attached adapters.
}
\vspace{-0.3cm}
\label{tab:attach_ablation}
\resizebox{0.4\textwidth}{!}
    {
\begin{tabular}{r|c|c|c}
\toprule
{Location} & {FD $\downarrow$} & {CLIP\textsubscript{sim} $\uparrow$} & {BRISQUE $\uparrow$} \\
\midrule
All layers               & 0.165     & 0.33     & 35.00       \\
Non-CA layers & 0.158     & 0.34     & 37.00       \\
\rowcolor{lightblue}
CA layers    & \textbf{0.047} & \textbf{0.36} & \textbf{39.50} \\
\bottomrule
\end{tabular}
}
\end{table}

\vspace{-0.55cm}
\paragraph{Impact of orthogonal regularization (OR).}
Table \ref{tab:orthogonal_ablation} demonstrates the impact of orthogonal regularization (OR). Without OR, the FD score is 0.143, semantic similarity (CLIP\textsubscript{sim}) drops to 0.29, and image quality (BRISQUE) is 37.92. When OR is applied, performance improves significantly across all metrics, with a much lower FD score of 0.047, higher semantic similarity of 0.36, and improved image quality (BRISQUE = 39.50). This highlights the effectiveness of orthogonal regularization in reducing bias and improving overall performance.

\begin{table}[h]
    \centering
\caption{Orthogonal Ablation}
\vspace{-0.3cm}
\label{tab:orthogonal_ablation}
\resizebox{0.4\textwidth}{!}
    {
\begin{tabular}{r|c|c|c}
\toprule
{Location} & {FD $\downarrow$} & {CLIP\textsubscript{sim} $\uparrow$} & {BRISQUE $\uparrow$} \\
\midrule
Ours $w \backslash o$ OR  & 0.143 & 0.29 & 37.92 \\
\rowcolor{lightblue}
Ours $w \backslash$ OR   & \textbf{0.047} & \textbf{0.36} & \textbf{39.50} \\
\bottomrule
\end{tabular}
}
\end{table}



\section{Conclusion}
\vspace{-0.1cm}
In this paper, we propose FairGen, a plug-and-play method that learns attribute latent directions in a self-discovering manner. Our method can not only jointly debias multiple attributes in DMs, but also enables the generated images to follow a prescribed attribute distribution. It is lightweight and can be integrated with other DMs without retraining. Extensive experiments on debiasing gender, racial, and their intersectional biases show that our method outperforms previous SOTA by a large margin. We believe that our work marks a critical advancement in addressing harmful societal stereotypes within diffusion models, and it contributes to the ethical real-world applications of text-to-image DMs.
\section*{Acknowledgements}
This work was supported in part by the National Natural Science Foundation of China under Grant No. 62306261, and the Shun Hing Institute of Advanced Engineering (SHIAE) Fund under Grant No. 8115074.
{
    \small
    \bibliographystyle{ieeenat_fullname}
    \bibliography{main}
}

\clearpage
\setcounter{page}{1}
\maketitlesupplementary
\section{Implementation Details}
We use Stable Diffusion v2.1 for all methods. We employ the prompt template ``\textit{a photo of the face of a} \{occupation\}\textit{, a person}''. At inference time, for each bias, we generate 100 images per occupation across 100 occupations, resulting in a total of 10,000 images. We set $\eta = \alpha = 1$,  and train for 1000 iterations with a learning rate of 1e-5. For gender bias, we use the CelebA \citep{liu2015faceattributes} dataset to train a binary classifier with two categories:\{male,female\}. For racial bias, we use the FairFace \citep{fairface-model} dataset to train a classifier with the following four categories: \textcolor{red1}{WMELH}=\{White, Middle Eastern, Latino Hispanic\}, \textcolor{blue1}{Asian}=\{East Asian, Southeast Asian\}, Black, and \textcolor{brown}{Indian}. Please refer to supplementary for more details. We also conduct experiments with other versions of Stable Diffusion.

\section{Compared Methods}\label{compare_methods}

\vspace{0.05cm}
\noindent\textbf{Finetuning for Fairness (F4Fair)}~\citep{shen2024finetuningtexttoimagediffusionmodels} is a retraining-based approach with two main technical innovations: (1) a distributional alignment loss that aligns specific attributes of generated images to a user-defined target distribution, and (2) adjusted direct finetuning (adjusted DFT) of the diffusion model’s sampling process, which uses an adjusted gradient to directly optimize losses on generated images.

\vspace{0.05cm}
\noindent\textbf{Inclusive Text-to-Image GENeration (ITI-GEN)}~\citep{yu2020inclusive} enhances fairness in text-to-image synthesis by incorporating reference images. Instead of relying solely on text prompts, ITI-GEN leverages visual exemplars to more effectively represent attributes that are difficult to describe in words, such as nuanced variations in skin tones. The key idea is to learn prompt embeddings that guide the generation process, ensuring balanced and inclusive outputs across different attribute categories.

\vspace{0.05cm}
\noindent\textbf{H-Distribution Guidance (H Guidance)}~\citep{parihar2024balancingactdistributionguideddebiasing} does not require retraining DMs. It introduces \textit{Distribution Guidance}, which ensures that generated images follow a prescribed attribute distribution. This is achieved by leveraging the latent features of the denoising UNet, which contain rich demographic semantics, to guide debiased generation. They also train an \textit{Attribute Distribution Predictor} (ADP), a small MLP that maps latent features to attribute distributions. ADP is trained using pseudo labels generated by existing attribute classifiers, allowing fairer generation with the proposed Distribution Guidance.

\vspace{0.05cm}
\noindent\textbf{Unified Concept Editing (UCE)}~\citep{gandikota2023unified} is a closed-form parameter-editing method that enables the application of numerous editorial modifications within a single text-to-image synthesis model, while maintaining the model’s generative quality for unedited concepts.

\vspace{0.05cm}
\noindent\textbf{Interpretable Diffusion}~\citep{li2024selfdiscoveringinterpretablediffusionlatent} is a self-supervised approach to find interpretable latent directions for a given concept. With the discovered vectors, it further propose a simple approach to mitigate inappropriate generation.


\section{More Visualization Results}
\label{appendix_results}
\begin{figure*} [tb!]
   \begin{center}
   \includegraphics[width=1.0\linewidth]{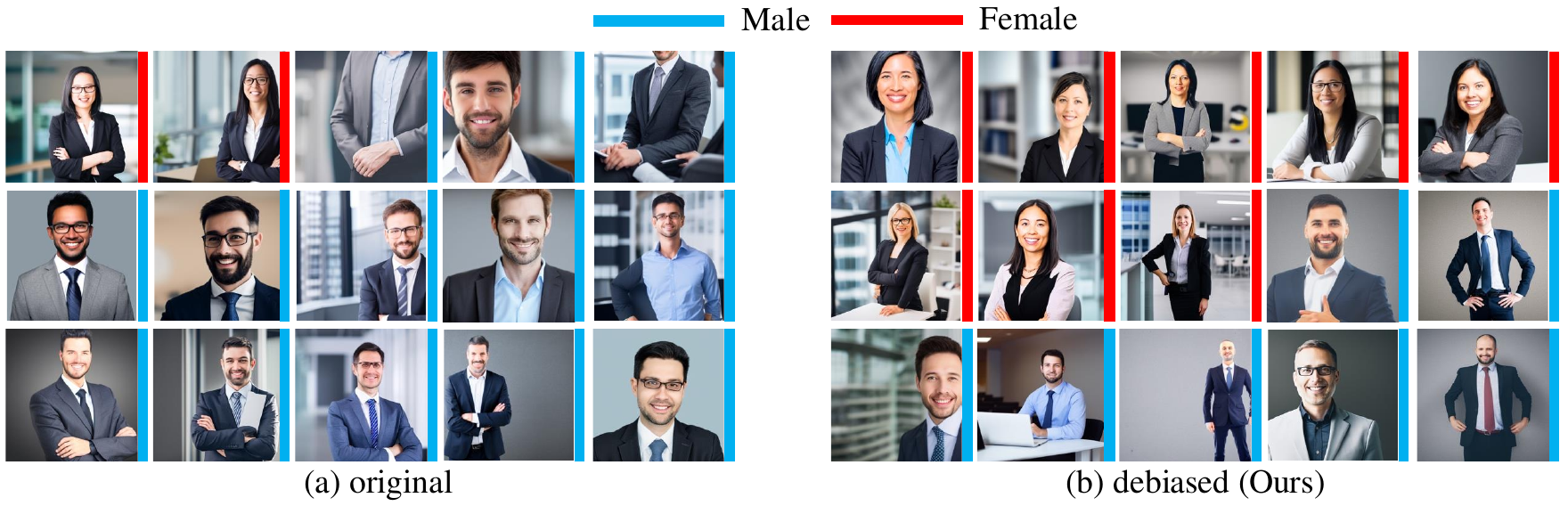} 
   \end{center} 
   \vspace{-0.6cm} 
      \caption{Images generated from the original SD (left) and Ours (right) for gender debias with prompt `A photo of a ceo'. Gendet ratio: Male : Female = 13 : 2 $\rightarrow$ 7 : 8}
\label{fig: appendix ceo gender debias}
   \end{figure*}

\begin{figure*} [tb!]
   \begin{center}
   \includegraphics[width=1.0\linewidth]{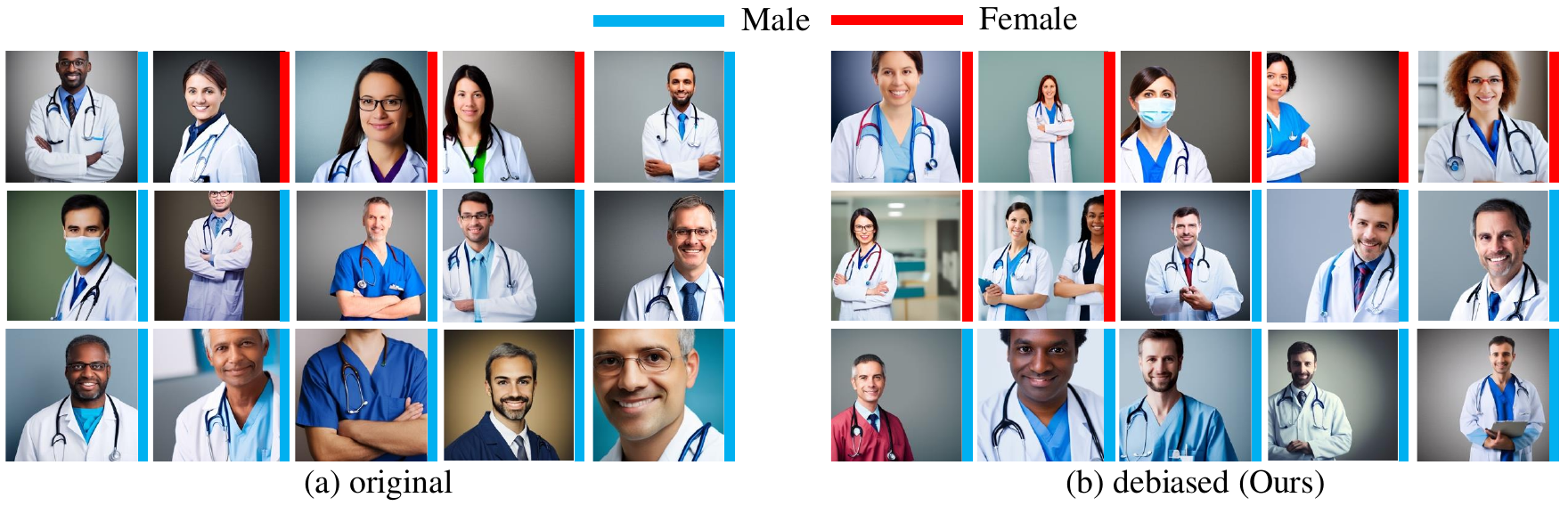} 
   \end{center} 
   \vspace{-0.6cm} 
      \caption{Images generated from the original SD (left) and Ours (right) for gender debias with prompt `A photo of a doctor'. Gendet ratio: Male : Female = 12 : 3 $\rightarrow$ 8 : 7}
\label{fig: appendix doctor gender debias}
   \end{figure*}

\begin{figure*} [tb!]
   \begin{center}
   \includegraphics[width=1.0\linewidth]{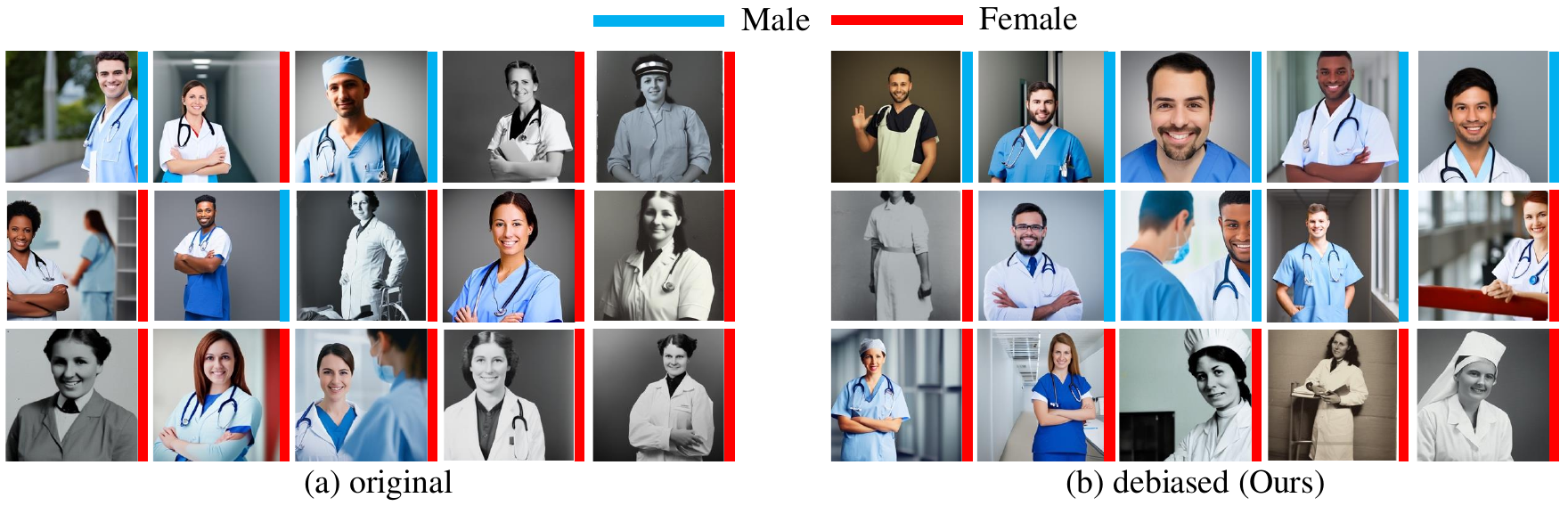} 
   \end{center} 
   \vspace{-0.6cm} 
      \caption{Images generated from the original SD (left) and Ours (right) for gender debias with prompt `A photo of a nusrse'. Gendet ratio: Male : Female = 3 : 12 $\rightarrow$ 7 : 8}
\label{fig: appendix nurse gender debias}
   \end{figure*}

We provide more visualization results about gender debaising and racial debaising. 
The qualitative results in Figure \ref{fig: appendix ceo gender debias} \ref{fig: appendix doctor gender debias} \ref{fig: appendix nurse gender debias} further demonstrate that our method(DebiasDiff) effectively mitigates gender bias without compromising image quality or semantic coherence.

\begin{figure*} [tb!]
   \begin{center}
   \includegraphics[width=1.0\linewidth]{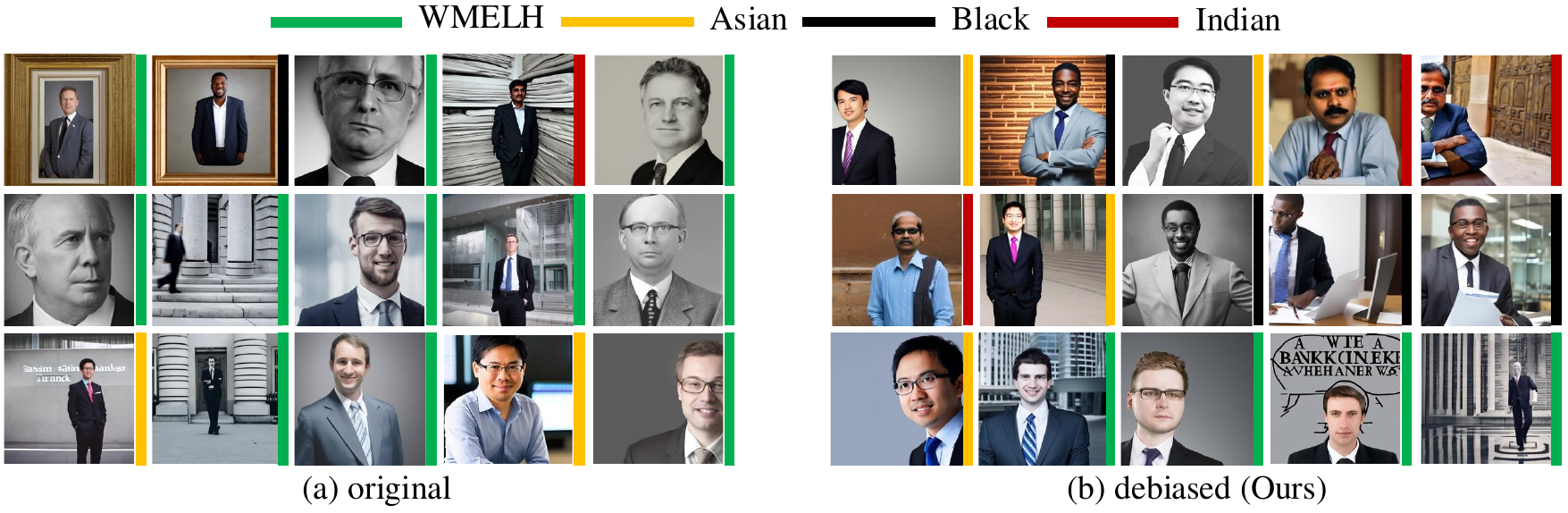} 
   \end{center}
   \vspace{-0.6cm} 
    \caption{Images generated from the original SD (left) and Ours (right) for race debias with prompt `A photo of a banker'. Racial group distribution: \textcolor{red1}{WMELH} : \textcolor{blue1}{Asian} : Black\textcolor{brown} :{Indian }= 10:2:1:1 $\rightarrow$ 4:4:4:3}
   \label{fig: appendix banker racial debias} 
   \end{figure*}

\begin{figure*} [tb!]
   \begin{center}
   \includegraphics[width=1.0\linewidth]{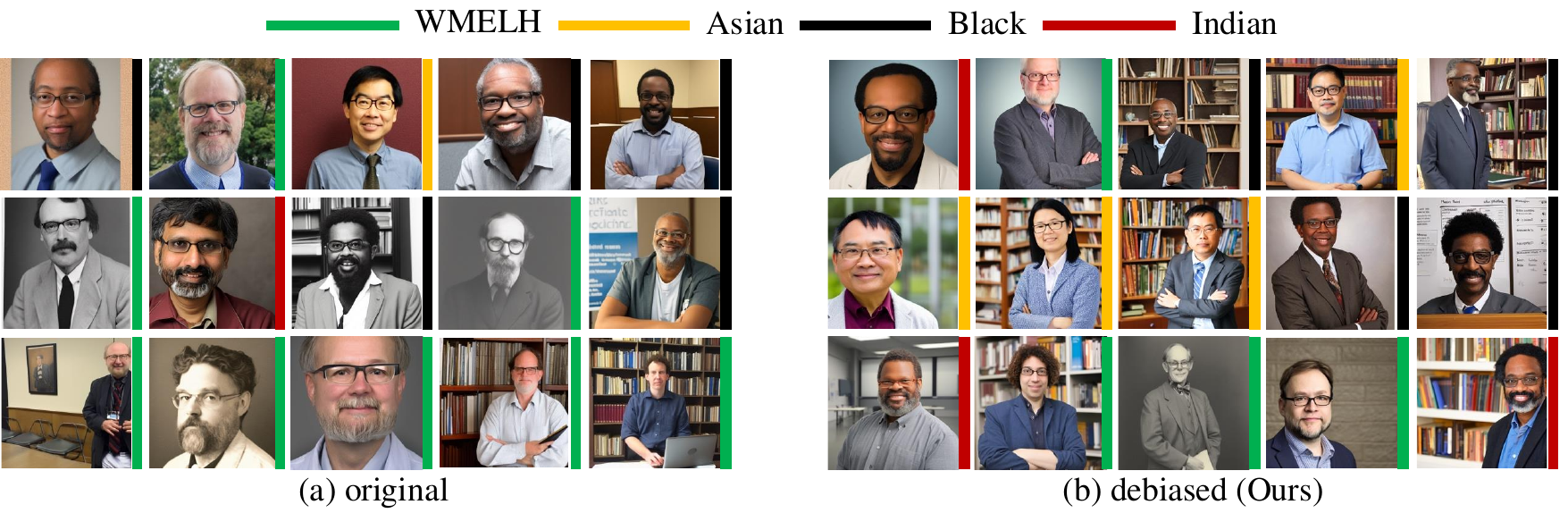} 
   \end{center}
   \vspace{-0.6cm} 
    \caption{Images generated from the original SD (left) and Ours (right) for race debias with prompt `A photo of a professor'. Racial group distribution: \textcolor{red1}{WMELH} : \textcolor{blue1}{Asian} : Black\textcolor{brown} :{Indian }= 8:1:5:1 $\rightarrow$ 4:4:4:3}
   \label{fig: appendix professor racial debias} 
   \end{figure*}

The qualitative results in  Figure \ref{fig: appendix banker racial debias} \ref{fig: appendix professor racial debias} further verify that our method outperforms others in reducing racial bias while preserving both semantic similarity and image quality.


\section{More Results on Scalability}

FairGen is designed with modularity and efficiency in mind, enabling scalable debiasing across multiple sensitive attributes. Its scalability stems from two core design choices: (1) a lightweight adapter architecture with linear complexity, and (2) an inference-time composition mechanism that avoids retraining or classifier dependency.

\vspace{-0.6cm}
\paragraph{Linear Adapter Complexity.}  
For \(k\) attributes with \(c\) categories each, FairGen introduces \(\mathcal{O}(kc)\) plug-and-play adapters. These adapters are trained independently and only the relevant ones are activated at inference time, ensuring that the computational cost remains bounded and practical, even as the number of attributes grows.

\vspace{-0.6cm}
\paragraph{Training Efficiency.}  
Each adapter is trained within 0.5 hours on a single A100 GPU. Even under multi-attribute settings, the total training time remains competitive with prior methods. As shown in Table~\ref{tab:efficiency}, FairGen achieves the lowest training time (1.0h) and identical inference speed (6.2s) to the base diffusion model, demonstrating its lightweight nature.

\begin{table}[h]
\caption{Training and inference efficiency comparison.}
\label{tab:efficiency}
\vspace{-0.85em}
\resizebox{0.48\textwidth}{!}{
\begin{tabular}{l|c|c|c|c|c|c|c}
\toprule
Metric & Original SD & F4Fair & ITI-GEN & H Guidance  & InterDiff & FairGen (Ours) \\
\midrule
Training Time  & -- & 4.3 h & 2.4 h & 2.8 h & 3.1 h  & \textbf{1.0 h} \\
Inference Time & \textbf{6.2s} & 6.8s & 6.4s & 7.1s & 6.9s  & \textbf{6.2s} \\
\bottomrule
\end{tabular}
}
\end{table}

\vspace{-0.6cm}
\paragraph{Inference Efficiency.}  
During inference, FairGen introduces negligible overhead by only activating the adapters corresponding to the target attribute categories. As Table~\ref{tab:efficiency} shows, its inference time matches that of the original Stable Diffusion, while achieving superior fairness.

\vspace{-0.6cm}
\paragraph{Scalability in Attribute Space.}  
To evaluate FairGen’s robustness in higher-dimensional fairness settings, we extend it to debias across four attributes: gender, race, age (``young'', ``middle-aged'', ``old''), and body type (``thin'', ``medium'', ``obese''). Table~\ref{tab:scalability} shows that FairGen consistently achieves low fairness discrepancy (FD) and maintains visual quality (CLIP\textsubscript{sim}, FID, BRIS) with only minor degradation, confirming its practical scalability.

\begin{table}[h]
\centering
\caption{Scalability analysis of FairGen on four attributes.}
\label{tab:scalability}
\vspace{-0.85em}
\resizebox{0.4\textwidth}{!}{
\begin{tabular}{l|c|c|c|c}
\toprule
Setting & FD $\downarrow$ & CLIP\textsubscript{sim} $\uparrow$ & FID $\downarrow$ & BRIS $\uparrow$ \\
\midrule
Gender & 0.041 & 0.37 & 12.34 & 38.52 \\
Gender+Race & 0.042 & 0.37 & 13.18 & 38.33 \\
Gender+Race+Age & 0.044 & 0.36 & 13.95 & 38.41 \\
Gender+Race+Age+Body Type & 0.045 & 0.36 & 13.78 & 38.27 \\
\bottomrule
\end{tabular}
}
\end{table}

\vspace{-0.6cm}
\paragraph{Orthogonality Regularization.}  
FairGen applies orthogonality regularization to mitigate attribute interference during training. While helpful, we observe that strict orthogonality is not essential for strong debiasing. Due to the lightweight adapter structure, the additional cost introduced by this regularization remains minor. In future work, scalability can be further improved by applying orthogonality selectively, e.g., only between interfering attribute groups identified through data-driven analysis.

Overall, these results demonstrate FairGen's ability to scale to a broad range of fairness settings while maintaining efficiency and quality, making it a practical solution for large-scale fair image generation.

\end{document}